\documentclass[10pt,twocolumn,letterpaper]{article}

\usepackage{iccv}
\usepackage{times}
\usepackage{epsfig}
\usepackage{graphicx}
\usepackage{amsmath}
\usepackage{amssymb}

\usepackage[pagebackref=true,breaklinks=true,colorlinks,bookmarks=false]{hyperref}
\usepackage{url}

\usepackage{booktabs}
\usepackage{xspace}
\usepackage{multirow, makecell}
\usepackage{color, colortbl}
\usepackage{graphicx}
\usepackage{wrapfig}
\usepackage{mathtools}
\usepackage{caption}
\usepackage{subcaption}

\usepackage{array}
\usepackage{multirow,multicol}
\usepackage[flushleft]{threeparttable}
\usepackage{comment}
\usepackage{algorithmic}
\usepackage{booktabs}
\usepackage{bbm, dsfont}
\usepackage[font=small,labelfont=bf]{caption}

\usepackage{amssymb}
\usepackage{pifont}

\newcommand{\tablestyle}[2]{\setlength{\tabcolsep}{#1}\renewcommand{\arraystretch}{#2}\centering\footnotesize}

\newcolumntype{x}[1]{>{\centering\arraybackslash}p{#1pt}}
\newcommand{\app}{\raise.17ex\hbox{$\scriptstyle\sim$}}

\newlength\savewidth\newcommand\shline{\noalign{\global\savewidth\arrayrulewidth
  \global\arrayrulewidth 1pt}\hline\noalign{\global\arrayrulewidth\savewidth}}
\usepackage{color, colortbl}
\usepackage[dvipsnames]{xcolor}
\definecolor{tabhighlight}{HTML}{e5e5e5}
\makeatletter\renewcommand\paragraph{\@startsection{paragraph}{4}{\z@}
  {.5em \@plus1ex \@minus.2ex}{-.5em}{\normalfont\normalsize\bfseries}}\makeatother

\def\ie{\emph{i.e}\onedot}

\makeatletter
\def\@fnsymbol#1{\ensuremath{\ifcase#1\or \dagger\or \ddagger\or
   \mathsection\or \mathparagraph\or \|\or **\or \dagger\dagger
   \or \ddagger\ddagger \else\@ctrerr\fi}}
\makeatother
\def\tablecite#1#{%
  \def\pretablecite{#1}%
  \tableciteaux}
\def\tableciteaux#1{%
  \textsuperscript{\expandafter\originalcite\pretablecite{#1}}%
}
\usepackage{graphicx}
\usepackage{enumitem}
\usepackage{wrapfig}
\usepackage{lipsum}
\usepackage{soul}

\newcolumntype{H}{>{\setbox0=\hbox\bgroup}c<{\egroup}@{}}
\newcolumntype{a}{>{\columncolor{Gray}}c}

\usepackage{tabu}
\usepackage{xcolor}
\usepackage{nicematrix}
\usepackage[accsupp]{axessibility}
\definecolor{ForestGreen}{rgb}{0.13, 0.55, 0.13}
\definecolor{Green}{rgb}{0.0, 0.5, 0.0}
\definecolor{green(munsell)}{rgb}{0.0, 0.66, 0.47}
\definecolor{green(ryb)}{rgb}{0.4, 0.69, 0.2}
\definecolor{green(pigment)}{rgb}{0.0, 0.65, 0.31}
\definecolor{citecolor}{HTML}{0071bc}
\definecolor{GrayXMark}{gray}{0.7}

\usepackage{tabularx}
\usepackage[export]{adjustbox}

\iccvfinalcopy 

\def\iccvPaperID{2613} 

\ificcvfinal\fi

\begin{document}

\title{EdaDet: Open-Vocabulary Object Detection Using Early Dense Alignment}

\author{\textbf{
Cheng Shi, 
Sibei Yang\thanks{Corresponding author}}\\ 
 School of Information Science and Technology, ShanghaiTech University \\
{\tt\small \{shicheng2022,yangsb\}@shanghaitech.edu.cn}\\
\small{Project page:} \href{https://chengshiest.github.io/edadet/}{\small{https://chengshiest.github.io/edadet}}
}

\maketitle
\ificcvfinal\thispagestyle{empty}\fi

\begin{abstract}

Vision-language models such as CLIP have boosted the performance of open-vocabulary object detection, where the detector is trained on base categories but required to detect novel categories. 
Existing methods leverage CLIP's strong zero-shot recognition ability to align object-level embeddings with textual embeddings of categories. 
However, we observe that using CLIP for object-level alignment results in overfitting to base categories, \textit{i.e.}, novel categories most similar to base categories have particularly poor performance as they are recognized as similar base categories. 
In this paper, we first identify that the loss of critical fine-grained local image semantics hinders existing methods from attaining strong base-to-novel generalization. Then, we propose Early Dense Alignment (EDA) to bridge the gap between generalizable local semantics and object-level prediction. 
In EDA, we use object-level supervision to learn the dense-level rather than object-level alignment to maintain the local fine-grained semantics. 
Extensive experiments demonstrate our superior performance to competing approaches under the same strict setting and without using external training resources, \textit{i.e.}, improving the $+8.4$\% novel box AP50 on COCO and $+3.9$\% rare mask AP on LVIS. 

\vspace{-5mm}
\end{abstract}
\section{Introduction}

\begin{figure}[t]
    \includegraphics[width=1.00\linewidth]{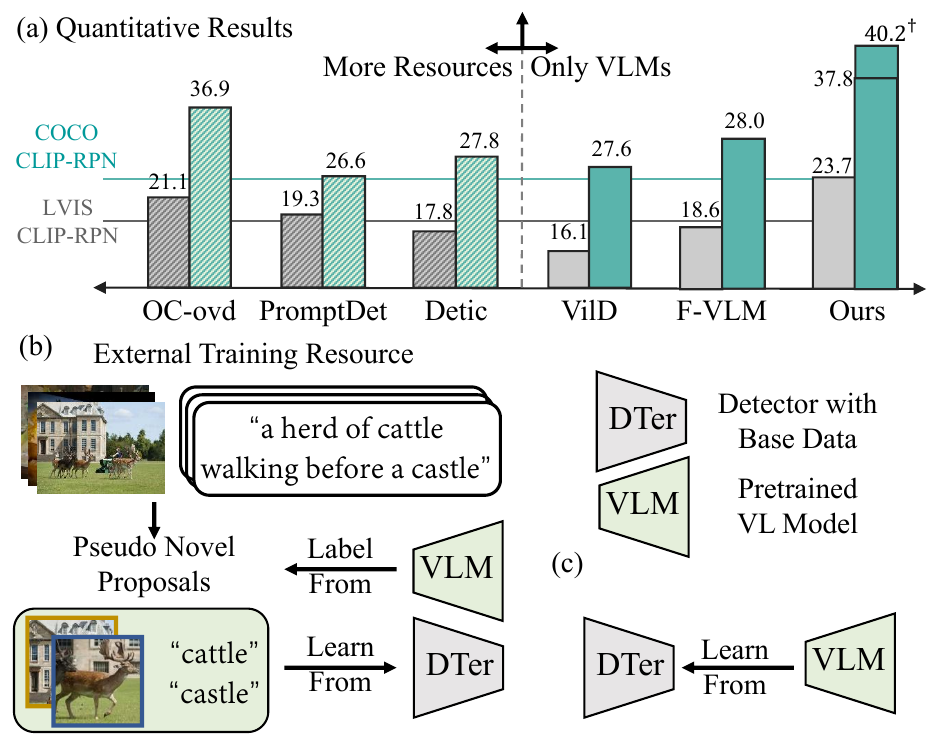}
    \vspace{-5mm}
    \caption{
    Different approaches to building an open-vocabulary detector: 
    (a) their performance comparison. ~$^\dagger$: with self-training. (b) generate pseudo ``novel" proposals from extra training resources and VLMs, or (c) generalize from VLMs
    }
    \label{fig:1}
    \vspace{-4mm}
\end{figure}

\begin{figure*}[t]
    \includegraphics[width=1.00\linewidth]{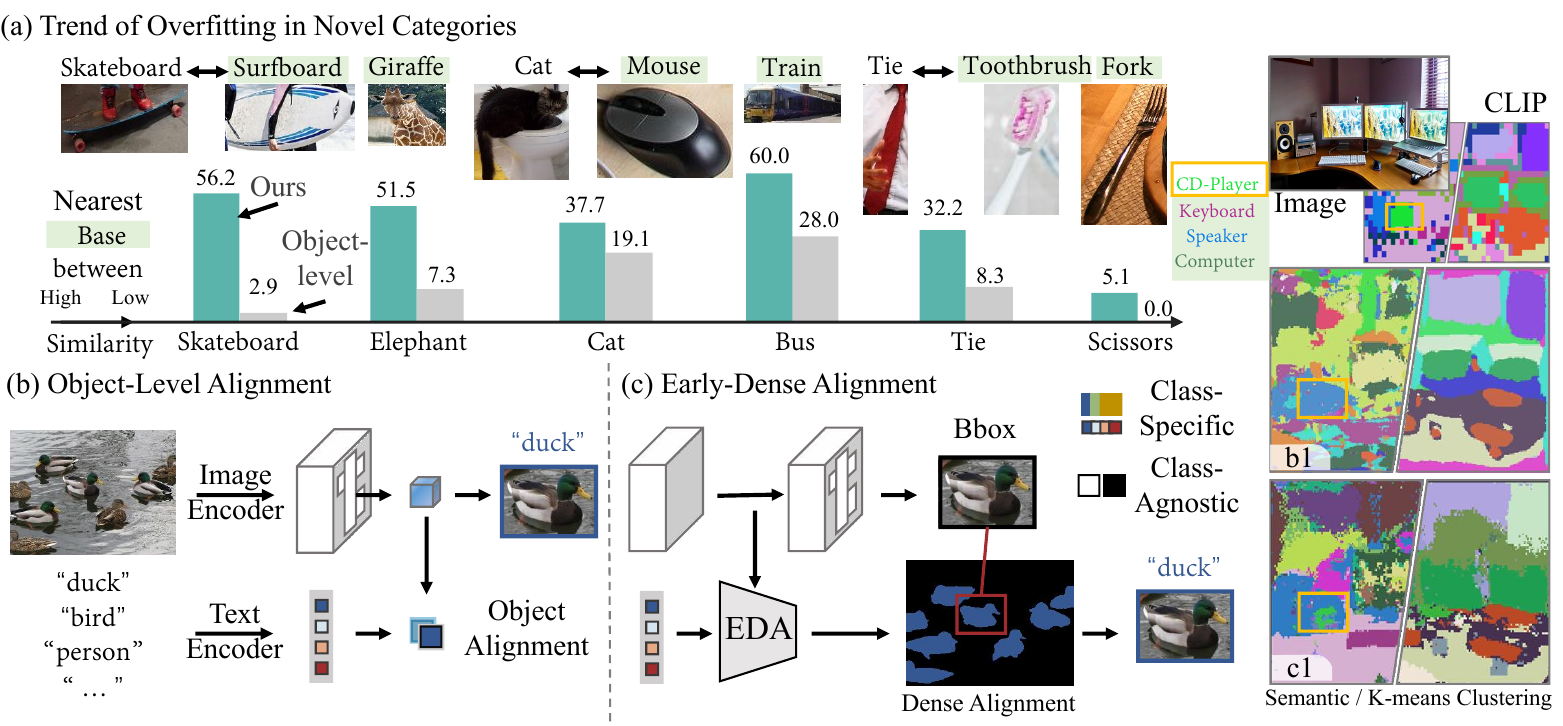}
    \caption{
    The comparison between Object-level Alignment and our Early Dense Alignment (Eda) on \textbf{(b)-(c)} architectures, \textbf{(b1)-(c1)} local image semantics and clustering results, and \textbf{(a)} 
    box AP of novel categories similar to base categories. We list six novel categories most similar to base categories by calculating the average similarity between the randomly sampled thousands of novel objects' visual features and base categories' text embeddings. Our Eda: (1) successfully recognizes the fine-grained \textit{novel CD-player} that is predicted to \textit{base speaker} by object-level alignment; (2) better groups local image semantics into object regions compared with CLIP; (3) achieves a much higher 
    novel box AP for predicting novel objects similar to base objects, showing that Eda can distinguish fine-grained details of similar novel and base categories. In contrast, object-level alignment overfits base categories.
    }
    \label{fig:2}
    \vspace{-4mm}
\end{figure*}

Open-vocabulary object detection aims to localize and recognize objects of both \textit{base categories} and \textit{novel categories} when only the training labels on base categories are available. 
Beyond focusing on object detection on a closed set of categories~\cite{he2017mask,detr,yolov3,ssd,fcos}, open-vocabulary detection requires generalizing well from base to all novel categories without annotations for each novel category.

One straightforward idea is to generate pseudo-proposals relevant to novel categories and train detectors with base and novel categories (see Figure~\ref{fig:1}\textcolor{red}{b}), adopted by~\cite{detic,vldet,promptdet,gap,locov}. They usually first extract concepts relevant to novel categories and then generate proposals of novel concepts from extra training resources. Among them, some works~\cite{detic,promptdet,gap} follow the weak open-vocabulary setting~\cite{ovrcnn}, where the class names of novel categories directly corresponding to novel concepts are available in the training phase. Alternatively, some works~\cite{regionclip,vldet} extract novel concepts from captions or image-text pairs. Although these approaches have improved the performance of detecting novel categories, the need for additional training resources that heavily overlap with or are relevant to novel categories would limit them to practical applications.

Recently, contrastive pre-training of vision-language models (VLMs) like CLIP~\cite{clip} and ALIGN~\cite{align} have shown strong open-vocabulary image recognition ability. 
Some open-vocabulary detection works~\cite{vild,detpro,regionclip,fvlm,ovdetr} explore utilizing VLMs to learn transferable object representations. 
Figure~\ref{fig:2}\textcolor{red}{b} shows a high-level abstraction of these open-vocabulary detection frameworks. 
Although they plug well-designed methods to close the gap between visual representation learning for objects and images, they only achieve comparable or slightly better performance than CLIP-RPN\footnote{CLIP-RPN baseline simply utilizes CLIP to classify cropped proposals generated by region proposal network (RPN) trained on base categories.} on novel categories (right part of Figure~\ref{fig:1}\textcolor{red}{a}). 
We also follow the line of works that aim to generalize VLMs for object detection without generating pseudo-proposals relevant to novel categories (see Figure~\ref{fig:1}\textcolor{red}{c}). And we explore   
\textit{how better to utilize VLMs for base-to-novel generalization in open-vocabulary object detection.}

In this paper, we start by discovering and analyzing the respective advantages of VLMs and existing open-vocabulary detection frameworks for object detection. 
First, we observe that \textit{VLMs can predict local image semantics for novel categories while existing frameworks are easier to overfit base categories.} 
As shown in the ``Semantic" of Figure~\ref{fig:2} \textcolor{black} {(CLIP)}, CLIP successfully recognizes the novel local regions of the ``CD player", while the existing framework classifies the novel ``CD player" as the ``speaker" in base categories.
The reason is that VLMs may have seen fine-grained image-text pairs describing local semantics during training. In contrast, existing frameworks directly align the object representations to the classifier of base categories, which loses the fine-grained details that distinguish novel objects from their similar base objects. Without fine-grained details, the object-level representations of novel objects and their similar base objects are similar, resulting in them being classified into base categories. 

Figure~\ref{fig:2}\textcolor{red}{a} (marked in grey) 
shows that the object-level alignment's prediction accuracy for novel objects similar to base objects is much lower. 
Therefore, we propose to \textit{avoid direct object-level alignment and fully utilize VLMs' ability to distinguish fine-grained details for similar novel and base objects to preserve the recognition ability of novel categories.}

Second, we observe that \textit{the existing framework can better group local image semantics into object regions than VLMs.} The reason is that its object-level supervision for object representations generated from local semantics improves local semantic consistency to objects. 
As shown in the ``K-means clustering" of Figure~\ref{fig:2}\textcolor{red}{-b1} and \ref{fig:2} (CLIP), the existing framework groups the ```keyboard" well (marked in yellow), while the corresponding two separate ``keyboard" regions in CLIP's clustering map are mixed and indistinguishable. \textit{Therefore, we propose to adopt object-level supervision for the dense alignment of local image semantics.}

Based on the above discoveries, we propose a simple but effective solution, named early dense\footnote{``early'' means the use of features in the early stages of the backbone, and ``dense'' means per-pixel alignment to text for obtaining object-level prediction (Note that only object-level annotations are used).} alignment (Eda), to combine the strengths of VLMs and the existing frameworks. 
To avoid overfitting to base categories caused by object-level alignment, Eda directly predicts object categories from local image semantics to fully distinguish the fine-grained details of similar base objects and novel objects. To maintain the local semantics consistent for better grouping and localization, we use object-level supervision to learn the dense-level alignment. 
As shown in Figure~\ref{fig:2}\textcolor{red}{c}, Eda first aligns local image semantics to the CLIP's semantic space early and then predicts object-level labels based on the dense probabilities to categories. 
Our Eda enables dense-level alignment for local image semantics, which is much more generalizable than late object-level alignment to novel categories. Meanwhile, it can better group local semantics to object regions (see Figure~\ref{fig:2}\textcolor{red}{c1}). Also, Figure~\ref{fig:2}\textcolor{red}{a} (mark in green) shows that our Eda significantly improves the prediction accuracy for novel categories similar to base categories.

Finally, we propose EdaDet, a simple open-vocabulary detection framework by leveraging our early dense alignment (Eda). For object localization, we follow existing works~\cite{gap,detic,vild,fvlm} to learn class-agnostic object proposals. 
To ensure an efficient end-to-end localization and recognition framework, we adopt a query-based proposal generation method like DETR~\cite{detr} but revise it to be class-agnostic. 
For open-vocabulary recognition, we apply our generalizable Eda to predict the categories of class-agnostic proposals. In addition, for better generalization, EdaDet deeply decouples the object localization and recognition by separating the open-vocabulary classification branch from the class-agnostic proposal generation branch at a more shallow layer of the decoder.

To evaluate the effectiveness of our EdaDet, we conduct experiments on LVIS~\cite{lvis}, COCO~\cite{coco}, and Objects365~\cite{objects365} benchmarks. 
In summary, our main contributions are as follows,
\begin{itemize}
\setlength{\itemsep}{0pt}
\setlength{\parsep}{0pt}
\setlength{\parskip}{0pt}
	\item We propose a novel and effective early dense alignment (Eda) for base-to-novel generalization in object detection without knowing the class names of novel categories and using extra training resources.
	\item We propose an end-to-end EdaDet framework, which deeply decouples the object localization and recognition
by separating classification from the recognition at a more shallow layer of the decoder. 
	\item Despite being simple, EdaDet achieves strong quantitative results, outperforming state-of-the-art methods with the strict setting on COCO and LVIS by $5.8$\% box AP$50$ and $2.0$\% mask AP on novel categories respectively. Moreover, EdaDet shows striking cross-dataset transferable capability. 
        \item \textcolor{black}{EdaDet shows impressive qualitative predictions on local image semantics and demonstrates efficient and effective performance improvement when scaling the model size thanks to our generalizable Eda and deeply decoupled detection framework.}
\end{itemize}

\section{Related Work}

\textbf{Transferable Representation Learning} 
explores learning transferable representations from source tasks with large-scale data in the pre-training stage and then adapt the representations to a variety of target downstream tasks~\cite{bengiorep1,qinghuarep2}. 
Based on whether the data used in the pre-training is labeled or not, the pre-training can be divided into supervised and unsupervised. 
Supervised pre-training~\cite{imagenetpretain,metalearning} is commonly employed in computer vision community. For example, image classification on ImageNet~\cite{imagenet} is often used as the pre-training task for the downstream visual recognition tasks~\cite{he2017mask,detr,yolov3,ssd,fcos,unet,deeplab,fcn}. 
In contrast, unsupervised pre-training proposes self-supervised tasks for pre-training on unlabeled data, including the generative learning~\cite{gpt,gpt2,mae,bert} and contrastive learning~\cite{mocov2,dino,simclr,simsiam,clip}. 
The unsupervised pre-training enables learning generally transferable knowledge from many large-scale unlabeled website data~\cite{laion,cc3m}. Among them, the contrastive vision-language models (VLMs) train a dual-modality encoder on large-scale image-text pairs to learn transferable visual representations with text supervision.
In this paper, we aim to transfer the general knowledge to recognize open-vocabulary objects in images and therefore select the contrastive VLMs that can align pairs of image and text as our source models.

\textbf{Visual Recognition from Generalizable VLMs}. 
The contrastive VLMs such as CLIP~\cite{clip} and ALIGN~\cite{align} pre-trained on large-scale image-text pairs have shown transferability to various visual recognition tasks, such as image classification~\cite{zhou2022cocoop,coop}, semantic segmentation~\cite{denseclip,languageg}, and object detection~\cite{vild,detpro,regionclip,gap}. 
With a handcraft prompt, 
VLMs can extract the category's text embeddings as the classifier for images. 
For image classification, zero-shot VLMs have already demonstrated strong zero-shot classification performance on various image classification tasks. 
CoOp~\cite{coop} and CoCoOp~\cite{zhou2022cocoop} further model the context words of a prompt with learnable vectors, thereby eliminating the need for handcrafted design. 
Unlike image classification methods that can directly use the CLIP's image encoder, semantic segmentation approaches such as DenseCLIP~\cite{denseclip} and LSeg~\cite{languageg} learn a segmenter from scratch by using the CLIP's text encoder as a frozen classifier for dense features. 
MASKCLIP+~\cite{maskclip} achieves better segmentation results on unseen categories by generating pseudo labeling of unseen categories and performing self-training with pseudo labels. 
In open-vocabulary detection, previous works~\cite{vild,detpro,regionclip,gap} also explore distilling knowledge from CLIP to detectors. However, they can only achieve a comparable performance of novel categories with zero-shot CLIP predicted on cropped class-agnostic proposals. 

Therefore, we explore fully utilize CLIP's ability to distinguish fine-grained details for similar novel and base objects to preserve the recognition ability of novel categories.

\textbf{Open-Vocabulary Object Detection.} 
Depending on the availability of novel categories' vocabulary during training, open-vocabulary detection is divided into strict~\cite{vild} (unknown) and weak~\cite{detic} (known) settings. Under the weak open-vocabulary detection setting, one basic solution is to generate pseudo-proposals of novel categories. Previous works~\cite{detic,gap,promptdet} all leverage additional image-level data to generate novel pseudo-proposals and train detectors with both base and novel categories. 
For the strict setting, recent works~\cite{ovdetr,fvlm,detpro,vild} mainly focus on generalizing CLIP to detect objects of novel categories. 
They all learn class-agnostic proposals and then classify proposals by using category names' text embedding extracted from CLIP's text encoder as the classification weights but differ in adopting different strategies to improve the generalization performance.  
For example, DetPro~\cite{detpro} designs a learnable prompt token instead of a hand-craft prompt to achieve a better generalization performance. RegionCLIP~\cite{regionclip} develops a region-text pre-training strategy to obtain fine-grained alignment between image regions and textual concepts, which is more suitable for object-level prediction. 
Furthermore, some methods such as ZSD-YOLO~\cite{zsdyolo} and ViLD~\cite{vild} align the proposal representations to that extracted from the CLIP's image encoder.
We follow the strict open-vocabulary detection setting. Instead of aligning object representations to CLIP's semantic space and relying on object-level alignment like previous works, we propose to align local image semantics to CLIP's space at the dense level to mitigate the overfitting issue.
\section{Method}

Our study is a first attempt to recognize open-vocabulary objects by utilizing dense-level alignment of local image semantics to CLIP's semantic space. 
We start with a brief introduction of CLIP (see Section~\ref{preliminary}) and 
a simple but necessary modification to its image encoder for dense-level prediction. 
Next, we introduce our overall detection framework (EdaDet), an end-to-end query-based object detection architecture, in Section~\ref{sec:proposal}. 
Moreover, the class-agnostic proposal generation is described in this section. 
Finally, we present the open-vocabulary object classification implemented by our early dense alignment (Eda) in Section~\ref{sec:dense}. 

\subsection{Preliminary}
\label{preliminary}
\noindent\textbf{CLIP}~\cite{clip} is trained on large-scale image-text pairs by image-level contrastive learning. 
Specifically, CLIP has a pair of image encoder $f(\cdot)$ and text encoder $g(\cdot)$. The image encoder $f(\cdot)$ can be presented into two parts: a visual backbone (\eg, ResNet~\cite{he2016deep} or ViT~\cite{vit}) denoted as $f_\text{Backbone}(\cdot)$ and the global feature aggregation layer (\eg, the last global attention pooling layer for ResNet) denoted as $f_\text{G-Pooling}(\cdot)$. 
The global attention pooling layer $f_\text{G-Pooling}(\cdot)$ is a single layer of multi-head attention~\cite{attention} that takes the globally average-pooled feature as a class token $[\text{cls}]$ and concatenates it with patch tokens $[\text{patches}]$ flattened from outputs of $f_\text{Backbone}(\cdot)$ as inputs. 
Given a text $\mathcal{T}$ and an image $\mathcal{I}$, CLIP computes the similarity $\mathcal{S}$ between $\mathcal{T}$ and $\mathcal{I}$ by:
\begin{equation}
\mathcal{S}_\text{CLIP}= \cos(f_\text{G-Pooling}(f_\text{Backbone}(\mathcal{I}))_{[\text{cls}]}, g(\mathcal{T})),
\end{equation}
where 
$\cos(\cdot,\cdot)$ represents the cosine similarity, and class token's feature $f_\text{G-Pooling}(f_\text{Backbone}(\mathcal{I}))_{[\text{cls}]}$ is the global feature. 

\noindent\textbf{Modification of CLIP for Dense Prediction.} 
As CLIP models for image-level prediction, it is not trivial to extract local patch prediction (\ie, the actual similarities between patches $f_\text{Backbone}(\mathcal{I})$ and text $\mathcal{T}$) from CLIP. 
Similar to MaskCLIP~\cite{maskclip} to reformulate the value-embedding layer of $f_\text{G-Pooling}(\cdot)$, we retain the global pooling layer but additionally conditioned on a diagonal mask $\mathcal{M}$ that 
prevents information exchange between patches. The modified formulation for dense prediction is expressed as,
\begin{equation}
\mathcal{S}_\text{CLIP}= \cos(f_\text{G-Pooling}(f_\text{Backbone}(\mathcal{I}) |\mathcal{M} )_{[\text{patches}]}, g(\mathcal{T})). 
\end{equation}
We adopt the modified pooling layer 
to produce the fine-grained dense prediction as shown in Figure~\ref{fig:2} (CLIP).

\subsection{Open-Vocabulary Object Detector (EdaDet)}

\label{sec:proposal}

\noindent\textbf{Problem Setup.} We share the same strict problem setup following previous works~\cite{vild,fvlm,ovdetr,detpro,simplevit,vldet}. 
Given an image $\mathcal{I} \in \mathbb{R}^{H 
 \times W \times 3}$, the detector predicts a set of bounding boxes with categories. 
 It is trained with detection annotations of base categories $\mathbb{C}_\text{train}$ but needs to detect objects of both base and novel categories $\mathbb{C}_\text{target}$ where $\mathbb{C}_\text{target} \neq \mathbb{C}_\text{train}$. 
It means that the detector requires to be capable of localizing and recognizing objects belonging to novel categories $\mathbb{C}_\text{novel} = \mathbb{C}_\text{target}-\mathbb{C}_\text{train}$ that are not seen in training. In the strict problem setting, the vocabulary set of novel categories $\mathbb{C}_\text{novel}$ is unavailable in training. 
 
\begin{figure}[t]
    \includegraphics[width=1.00\linewidth]{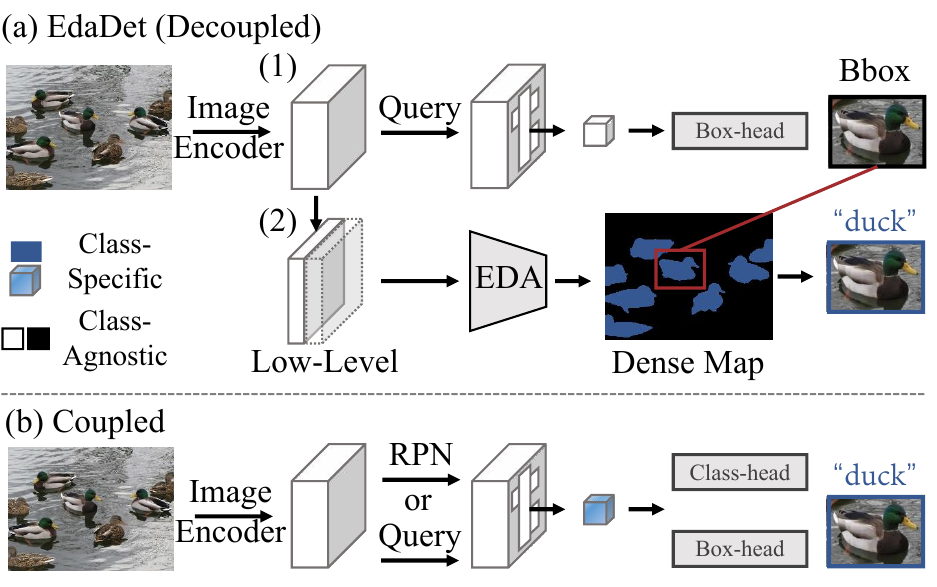}
    \caption{Architecture comparison between (a) our deeply-decoupled EdaDet and (b) existing open-vocabulary detection framework. EdaDet separates the open-vocabulary classification branch from the class-agnostic proposal generation branch at a more shallow layer of the decoder. EdaDet first individually generates object proposals and predicts dense probabilities to categories for local image semantics and then computes object proposals' categories based on the dense probabilities.}
    \label{fig:method}
    \vspace{-4mm}
\end{figure}

\noindent\textbf{The Overall EdaDet Architecture} is shown in Figure~\ref{fig:method}\textcolor{red}{a}. 
Following previous works~\cite{vild,detic,fvlm,gap}, we break down open-vocabulary detection as two subsequent branches: (1) to generate class-agnostic object proposals and (2) to recognize open-vocabulary categories for these object proposals. 

\label{sec:decouple}
For the proposal generation, to ensure an efficient end-to-end detector and avoid hand-designed modules like anchor generation, we adopt a query-based proposal generation method like DETR~\cite{detr} and retrofit it to class-agnostic. 
Specifically, for each object query, we predict a class-agnostic object confidence score $\mathcal{S}_\text{obj}$ to evaluate whether it is an object or background and regress its bounding box $\mathcal{B}$. We also use the bipartite matching loss for set prediction following DETR and denote the sum of confidence score loss and bounding box regression loss as $\mathcal{L}_{\text{box}}$. 
By the class-agnostic retrofit, the object proposals generated by our proposal generation branch have similar \textcolor{black}{top-$300$ average recall} to the RPN network of the existing Mask R-CNN detector.

However, we observe that both the Mask R-CNN detector and our initial attempt predict the bounding box regression on the same feature space as classifying open-vocabulary proposal with only using \textcolor{black}{two separate heads} for box regression and classification, as shown in Figure~\ref{fig:method}\textcolor{red}{b}. 
\textcolor{black}{We believe that the highly correlated predictions between the two branches hurt the class-agnostic proposal generation and validate this observation in Table~\ref{tab:proposal}.} 
Also, OLN~\cite{oln} shares a similar observation to us for learning open-world object proposals. 
Therefore, we separate the open-vocabulary classification from the proposal generation branch at a more shallow layer of the decoder, as shown in Figure~\ref{fig:method}\textcolor{red}{a}, to deeply decouple the two branches.

\subsection{Early Semantic Alignment at Dense Level (Eda)}
\label{sec:semantic}

Our early dense alignment (Eda) performs open-vocabulary classification for class-agnostic proposals by leveraging the generalizable CLIP. 
As in previous methods~\cite{regionclip,vild,fvlm,detic,gap,vldet}, we use text embeddings of categories extracted from frozen CLIP's text encoder $g(\cdot)$ as the classifier and denote the set of text embeddings of base categories $\mathbb{C}_\text{train}$ in training as $E_\text{train}$. 

However, unlike previous methods that classify proposals by aligning their object-level visual features to the base classifier, we first align dense local image semantics early to the classifier by using object-level supervision. Then, we classify proposals according to the dense probabilities to categories to mitigate the overfitting to base categories $\mathbb{C}_\text{train}$ led by object-level alignment. 
\textcolor{black}{Our dense-level alignment can persevere the fine-grained recognition ability to distinguish local semantics details for similar novel and base categories, which further helps to better generalize from similar base objects to novel objects.} 

\noindent\textbf{Early Dense Semantic Alignment.} 
Given the input image $\mathcal{I}$, we first extract its image feature map $\mathcal{F}_{i}(\mathcal{I})$ via our visual backbone $\mathcal{F}_{i}(\cdot)$, where feature at each spatial position of the feature map $\mathcal{F}_{i}(\mathcal{I})$ represents a local image semantic and $i$ represents the feature of the i-th layer. Then, we calculate the probability map of each local image semantic belonging to each base category as follows,
\begin{equation}
\mathcal{S}_\text{detector}= 
\text{Softmax}(\cos(\mathcal{F}_{i}(\mathcal{I}), E_\text{train})/\tau),
\label{equ:detector}
\end{equation}
where $\cos(\cdot,\cdot)$ represents the cosine similarity, and $\tau$ is the temperature coefficient. 
We also compute the CLIP's dense probability map $\mathcal{S}_\text{CLIP}(f_{\text{Backbone}}(\mathcal{I}), E_\text{train})$, \ie, $\text{Softmax}(\cos(f_{\text{G-Pooling}}(f_{\text{Backbone}}(\mathcal{I})|\mathcal{M})_{[\text{patches}]}, E_\text{class})/\tau)$. 
Inspired by F-VLM~\cite{fvlm} that fuse objects' CLIP scores and detection scores via geometric mean, we obtain the overall dense score map $\mathcal{S}$ as follows,
\begin{equation}
\mathcal{S} =  \mathcal{S}_\text{detector}^{1-\lambda} \circ \mathcal{S}_\text{CLIP}^\lambda,
\label{equ:detector_all}
\end{equation}
where $\lambda \in [0,1]$ controls weights for the probability maps $\mathcal{S}_\text{detector}$ and $\mathcal{S}_\text{CLIP}$, and $\circ$ means element-wise product. 

Next, we classify class-agnostic proposals generated from our proposal generation branch based on the overall dense score map $\mathcal{S}$. Given a proposal with predicted bounding box $\mathcal{B}$, we pool the box $\mathcal{B}$ into a fixed size score map $\text{RoIAlign}(\mathcal{B}, \mathcal{S})$ by performing RoIAlign~\cite{he2017mask} operation on the overall scores $\mathcal{S}$. 
Based on our observation that an object typically does not occupy the entire box, we independently average the top-$k$ highest scores for each category in the fixed size score map $\text{RoIAlign}(\mathcal{B}, \mathcal{S})$ as the proposal's predicted score for that category, which is formulated as:
\begin{equation}
\mathcal{S}_{\text{proposal}} = \text{M-Pooling}(\text{RoIAlign}(\mathcal{B}, \mathcal{S}) \circ \mathcal{M}_k),
\label{equ:sample}
\end{equation}
where $\text{M-Pooling}(\cdot)$ means the mean pooling layer. And mask $\mathcal{M}_k(i,j,c)=1$ means that the score $\text{RoIAlign}(\mathcal{B}, \mathcal{S})(i,j,c)$ at spatial position $(i,j)$ is among the top-$k$ highest scores for the $c$-th category, and otherwise $\mathcal{M}_k(i,j,c)=0$. And $\mathcal{S}_\text{proposal}$ represent the proposal's scores to each base category. \textcolor{black}{Note that the mask $\mathcal{M}_k$ allows non-object parts to retain their original semantics and avoid their overfitting to the proposal's category.}  

During training, we minimize the cross-entropy loss $\mathcal{L}_\text{cls}$ with respect to each proposal's classification scores $\mathcal{S}_\text{proposal}$. During inference, we replace the text embeddings of base categories with that of target categories and predict the proposal's label as the category with the highest score. 

\label{sec:dense}
\noindent\textbf{Global Semantic Alignment.} 
We observe that only local alignment leads to losing the global semantic information for the image. Therefore, we align the integrated local image semantics to CLIP's image encoder to improve the dense alignment. 
Specifically, given the global image feature $f_\text{G-Pooling}(f_\text{Backbone}(\mathcal{I}))_{[\text{cls}]}$ extracted from CLIP, we apply $L_1$ loss to align the local image semantics $\mathcal{F}(\mathcal{I})$:
\begin{equation}
\mathcal{L}_\text{g} = \|\text{M-Pooling}(\mathcal{F}(\mathcal{I}))-f_\text{G-Pooling}(f_\text{Backbone}(\mathcal{I}))_{[\text{cls}]}\|_{1}.
\label{equ:global}
\end{equation}

\section{Experiment}
\begin{table*}[t]
\tablestyle{3pt}{1.0}
\small
\begin{center}
\begin{tabular}{p{3.1cm}cc>{\color{gray}}c>{\color{gray}}clc>{\color{gray}}c>{\color{gray}}c>{\color{gray}}cc}
\multirow{2}{*}{Methods} & \multirow{2}{*}{\makecell{Publication}} & \multirow{2}{*}{\makecell{Training Resources for\\ Novel Proposal Generation}} & \multicolumn{3}{c}{{COCO Detection}} && \multicolumn{4}{c}{{LVIS Segmentation}} \\
\cline{4-6} \cline{8-11}
& & & \multicolumn{1}{c}{AP50$^{\text{box}}$} & \multicolumn{1}{c}{AP50$^{\text{box}}_{\text{base}}$} & \multicolumn{1}{c}{AP50$^{\text{box}}_{\text{novel}}$} 
&&   \multicolumn{1}{c}{AP$^{\text{mask}}$} & \multicolumn{1}{c}{AP$^{\text{mask}}_{\text{f}}$} & \multicolumn{1}{c}{AP$^{\text{mask}}_{\text{c}}$} & \multicolumn{1}{c}{AP$^{\text{mask}}_{\text{novel}}$}  \\ [.1em]
\shline
\multicolumn{4}{l}{\bf \textit{Weak open-vocabulary Setting}} &&&&& \\

Detic~\cite{detic} & ECCV2022 & IN-O &  \phantom{1}45.0 &  \phantom{1}47.1 &  \phantom{1}27.8 &&  \phantom{1}26.8 &  \phantom{1}31.6 &  \phantom{1}26.3 &  \phantom{1}17.8 \\



PromptDet~\cite{promptdet} & ECCV2022 & LAION &   \phantom{1}50.6 &  \phantom{1}- &  \phantom{1}26.6  &&  \phantom{1}21.4 &  \phantom{1}25.8 &  \phantom{1}18.3 &  \phantom{1}19.3  \\

VL-PLM~\cite{vlplm} & ECCV2022 & - &   \phantom{1}48.3 &  \phantom{1}54.0 &  \phantom{1}32.3  &&  \phantom{1}- &  \phantom{1}- &  \phantom{1}- &  \phantom{1}-  \\

PB-OVD~\cite{pbovd} & ECCV2022 & CAP-L &   \phantom{1}42.1 &  \phantom{1}46.1 &  \phantom{1}30.8  &&  \phantom{1}- &  \phantom{1}- &  \phantom{1}- &  \phantom{1}-  \\


OC-ovd~\cite{gap} & NeurIPS2022 & IN-O + COCO CAP& \phantom{1}51.5 &   \phantom{1}56.6 &  \phantom{1}36.9  &&  \phantom{1}25.9 &  \phantom{1}29.1 &  \phantom{1}25.0 &  \phantom{1}21.1  \\

LocOv~\cite{locov} & GCPR2022 & COCO CAP&   \phantom{1}45.7 &  \phantom{1}51.3 &  \phantom{1}28.6  && \phantom{1}- &  \phantom{1}- &  \phantom{1}- &  \phantom{1}-  \\

\hline
\multicolumn{4}{l}{\bf \textit{Strict open-vocabulary Setting}} &&&&& \\

CLIP-RPN & - & - &    \phantom{1}27.8 &  \phantom{1}28.3 &  \phantom{1}\underline{26.3}  &&   \phantom{1}17.7 &  \phantom{1}16.0 &  \phantom{1}18.8 &  \phantom{1}\underline{18.9}  \\

OVR-CNN~\cite{ovrcnn} & CVPR2021 & COCO CAP & \phantom{1}39.9 &  \phantom{1}46.0 &
\phantom{1}22.8
&&   \phantom{1}- &  \phantom{1}- &  \phantom{1}- &  \phantom{1}-  \\

ViLD~\cite{vild} & ICLR2022 & - &    \phantom{1}51.3 &  \phantom{1}59.5 &  \phantom{1}27.6  &&  \phantom{1}22.5 &  \phantom{1}28.3 &  \phantom{1}20.0 &  \phantom{1}16.1  \\
ViLD-Ens.~\cite{vild} & ICLR2022 & -&  \phantom{1}- &  \phantom{1}- &  \phantom{1}-  && \phantom{1}25.5 &  \phantom{1}30.3 &  \phantom{1}24.6 &  \phantom{1}16.6  \\

DetPro~\cite{detpro} & CVPR2022 & - &   \phantom{1}- &  \phantom{1}- &  \phantom{1}- &&  \phantom{1}25.9 &  \phantom{1}28.9 & \phantom{1}25.6 &  \phantom{1}19.8 \\

RegionCLIP~\cite{regionclip} & CVPR2022 & COCO CAP + CC3M&    \phantom{1}50.4 &  \phantom{1}57.1 &  \phantom{1}31.4  &&   \phantom{1}28.2 &  \phantom{1}34.0 &  \phantom{1}27.4 &  \phantom{1}17.1  \\



OV-DETR~\cite{ovdetr} & ECCV2022 & -&   \phantom{1}52.7 &  \phantom{1}61.0 &  \phantom{1}29.4  &&  \phantom{1}26.6 &  \phantom{1}32.5 &  \phantom{1}25.0 &  \phantom{1}17.4  \\

OWL-ViT~$^\dagger$~\cite{gap} & ECCV2022 & - &    \phantom{1}- &  \phantom{1}- &  \phantom{1}-  &&  \phantom{1}19.3 &  \phantom{1}- &  \phantom{1}- &  \phantom{1}16.9  \\

VLDet~\cite{vldet} & ICLR2023 & COCO CAP + CC3M&   \phantom{1}45.8 &  \phantom{1}50.6 &  \phantom{1}32.0  &&  \phantom{1}30.1 &  \phantom{1}34.3 &  \phantom{1}29.8 &  \phantom{1}21.7  \\

F-VLM~\cite{fvlm} & ICLR2023 & - &    \phantom{1}39.6 &  \phantom{1}- &  \phantom{1}28.0  &&  \phantom{1}24.2 &  \phantom{1}26.9 &  \phantom{1}24.0 &  \phantom{1}18.6  \\
\rowcolor{tabhighlight}
 Ours (RN50 backbone) & ICCV2023 & - &  \phantom{1}52.5	&	\phantom{1}57.7	&	\phantom{1}\textbf{37.8} && \phantom{1}27.5	&	\phantom{1}29.1	&	\phantom{1}27.5 & \phantom{1}\textbf{23.7} 	\\
\hline
\end{tabular}
\end{center}
\vspace{-14pt}
\caption{\textbf{Open-vocabulary object detection results on COCO and LVIS datasets.} All the methods share the RN50 backbone except $^\dagger$ with ViT-B/32. The IN-O is the ImageNet21k~\cite{imagenet} (IN-21K) dataset's subset with 997 overlapping categories with LVIS~\cite{lvis}. LAION~\cite{laion} is a large-scale image-text dataset, CC3M represents the Conceptual Caption 3M~\cite{cc3m}, and CAP-L consists of COCO Caption~\cite{coco}, Visual-Genome~\cite{visual} and SBU Caption~\cite{sbu}.
}
\vspace{-16pt}
\label{tab:1}
\end{table*}

\subsection{Benchmark and Implementation Detail}

\noindent\textbf{LVIS benchmark}~\cite{lvis} consists of object detection and instance segmentation labels for 1203 object categories. The categories are split into three groups: frequent, common, and rare. 
Following ViLD~\cite{vild}, we treat frequent and common categories as the base categories during training and treat the 337 rare categories as the novel categories during testing. And the 337 rare categories are excluded from the training set. 
The mask mAP on novel categories is the key evaluation metric for LVIS. 

\noindent\textbf{COCO benchmark}~\cite{coco} is a common benchmark for numerous studies on open-vocabulary detection. The COCO vocabulary is partitioned into 48 base categories for training and 17 novel categories for testing. We conform to the standard protocol and report results under the generalized detection settings. 
The key evaluation metric is the box AP50 of novel categories.

\noindent\textbf{Implementation Details.} Our detection framework is based on DETR~\cite{detr} following OV-DETR~\cite{ovdetr}. For fair comparison, we follow~\cite{detic,ovdetr} to train our detector for 10.2k iterations with batch size $32$ and image size 800 $\times$ 800 and adopt AdamW optimizer with weight decay 1e-4 and initial learning rate 2e-4. We set the fixed size of the score map and $k$ in Eq~\ref{equ:sample} as 14$\times$14 and 12$\times$12, respectively. The temperature coefficient $\tau$=$1$e$3$ in Eq~\ref{equ:detector}, and $\lambda$ = $0.25$ in Eq~\ref{equ:detector_all}. 
We use standard CLIP's ImageNet prompts to extract text embeddings by CLIP-R50. 
We fuse feature maps from conv2 x and conv3 from the backbone to perform Early Dense Alignment, i.e. $i$=$2$,$3$ in Equ~\ref{equ:detector}.
Unless otherwise specified, the experiments are conducted under the same setting.


\subsection{Comparison with State-of-the-Art Detectors}
\label{sec:mainresults}
We evaluate EdaDet on various open-vocabulary object detection and instance segmentation benchmarks.  
Results in Table~\ref{tab:1} show that EdaDet achieves a stronger base-to-new generalization on both COCO and LVIS benchmarks. 
Despite being trained in a strict open-vocabulary setting and without using any additional training resources, EdaDet consistently outperforms all prior methods, even these under more relaxed experimental setups.  


\noindent\textbf{COCO Benchmark.} Table~\ref{tab:1} and Table~\ref{tab:ablations} (w self-training) show that 
EdaDet consistently and significantly outperforms state-of-the-art methods under both weak and strict open-vocabulary settings. 
Specifically, EdaDet outperforms OV-DETR~\cite{ovdetr}, which shares the same setting (strict and without using any additional training resources) and detection framework with us by $\mathbf{+8.4}$ box AP50 of novel categories. 
\textcolor{black}{Compared to the strict best-performing VLDet~\cite{vldet} that is trained with additional caption resources, we improve box AP50 of novel categories by $\mathbf{+5.8}$.} 
Moreover, following the same weak open-vocabulary setting as best-performing OC-ovd~\cite{gap}, our EdaDet (without using external training resources) can further boost the box AP50 of all categories and the novel categories to $57.1$ and $40.2$ (see Table~\ref{tab:ablations}), which outperforms all state-of-the-art methods by $\mathbf{+5.6}$ and $\mathbf{+3.3}$, respectively. Different from LVIS, we observed significant overfitting to the base classes on the COCO dataset. Therefore, for COCO, we took the specific approach of using an RPN trained on the base classes to additionally extract class-agnostic proposals.

\noindent\textbf{LVIS Benchmark.} Table~\ref{tab:1} demonstrates that EdaDet achieves the new state-of-the-art mask AP50 of novel categories and is very competitive under different experimental settings. Compared to approaches under the same setting with us, EdaDet offers better performance, \ie, $\mathbf{+3.9}$ mask AP$_{\text{Novel}}$ improvement compared with best-performing DetPro~\cite{detpro}. 
Compared to the leading method VLDet~\cite{vldet} which leverages a large image-text dataset CC3M to expand vocabulary, our method directly transfers knowledge from VLMs and improves the mask AP$_{\text{Novel}}$ by $\mathbf{+2.0}$. Moreover, in contrast with ViLD~\cite{vild}, our EdaDet still preserves the performance of
base categories when improving the novel classes. Even compared with the ensemble version of ViLD-ens, EdaDet still boosts the performance by $\mathbf{+2.0}$ and $\mathbf{+7.1}$ on mask AP$_{\text{All}}$ and mask AP$_{\text{Novel}}$, respectively.

\noindent\textbf{Transfer Detection Benchmark.} We conduct a transfer detection experiment to assess the effectiveness of EdaDet as a universal detector for various data sources. 
Considering the base categories of the LVIS dataset contain almost all the COCO categories, and COCO shares the same image sources as LVIS, we do not conduct the transfer detection from LVIS to COCO like ViLD~\cite{vild}. Instead, following~\cite{vldet}, we conduct a more challenging transfer experiment that evaluates the COCO-trained model on LVIS and the LVIS-trained model on Objects365~\cite{objects365}. 
Specifically, we simply replace the COCO-based classifier (80 categories) with the LVIS-based classifier (1203 categories) without fintuning and report the box AP as the evaluation metric on LVIS of all categories. The transfer from LVIS to Objects365 follows a similar protocol. 
The results are shown in Table~\ref{table:transfer}.
For the transfer from COCO to LVIS, we observe that the method~\cite{vldet} under strict open-vocabulary setting usually outperforms those under weak setting~\cite{gap,detic} due to the better generalization ability of the former. 
Our EdaDet outperforms the best-performing strict VLDet~\cite{vldet} by $\mathbf{+2.8}$ box AP$_{50}$.
For objects365, EdaDet outperforms state-of-the-art methods~\cite{fvlm, detpro, vild} by $\mathbf{+1.7}$ and $\mathbf{+2.0}$ in terms of AP and AP$_{75}$, respectively.  

\begin{table}[t]
\small
\tabcolsep=0.5em
\centering
    \begin{tabular}{l|ccc|l|ccc}
    \toprule
      \multirowcell{2}[0ex][l]{Method} & \multicolumn{3}{c|}{COCO $\rightarrow$ LVIS} & & \multicolumn{3}{c}{LVIS $\rightarrow$ Objects365}\\
    & AP & AP$_{50}$ & AP$_{75}$ & & AP & AP$_{50}$ & AP$_{75}$ \\
    \midrule
    OC-ovd$^\dagger$ & 5.6 & 8.5 & 6.0 &VilD& 11.8& 18.2 & 12.6 \\
     Detic$^\dagger$      &  5.5 & 8.5 & 5.8 &DetPro& 12.1 & 18.8 & 12.9 \\
     VLDet      & - & 10.0 & - &F-VLM& 11.9 & 19.2 & 12.6 \\
     \rowcolor{tabhighlight}
     Ours    &  \textbf{9.1} & \textbf{12.8} & \textbf{9.6} &Ours& \textbf{13.6} & \textbf{19.8} & \textbf{14.6} \\
    \bottomrule
    \end{tabular}
    \caption{\textbf{Transfer detection of EdaDet}. We evaluate COCO-trained model on LVIS and LVIS-trained model on Objects365 without finetuning. $^\dagger$: evaluate with official code and checkpoint.}
    \vspace{-5mm}
    \label{table:transfer}
\end{table}

\begin{table}[t]
\small
\centering
    \begin{tabular}{l|c|c|cc|c}
    \toprule
    Method & Size & AP$^{\text{mask}}_{\text{novel}}$ & \#Iters & Epochs \\
    \midrule
    ViLD-EN-B7 & - & 26.3 & 180k & 460  \\
    \midrule
    OWL-ViT-Large & 1216 & 31.2 & 70k & 180 \\
    F-VLM-RN50x64 & 812 & 32.8 & 46.1k & 118   \\
    VLDet-ViT-Base & 345 & 26.3 & 90k & 13   \\
    \rowcolor{tabhighlight}
     EdaDet-ViT-Base & 345 & 29.9 & 42k & 10  \\
     \rowcolor{tabhighlight}
     EdaDet-ViT-Base & 345 & 35.6 & 126k & 30  \\
    \bottomrule
    \end{tabular}
    \caption{\textbf{Comparison of scalability and training efficiency} on LVIS. We report AP$^{\text{mask}}_{\text{novel}}$ to show the trade-off between performance and training costs under relatively fair model sizes in MB.}
    \vspace{-5mm}
    \label{table:compute}
\end{table}

\noindent\textbf{Scalability and Training Efficiency Benchmark.} Table~\ref{table:compute} summarizes the performance, iteration and epoch of different methods with the strict setting using large backbone networks. 
Notice that since different methods adopt different backbones and batch sizes, Table~\ref{table:compute} is not for an apple-to-apple comparison but just to illustrate the scalability and training efficiency for different methods. \textcolor{black}{With a relatively smaller backbone ($1/4$ of OWL-ViT), EdaDet achieves a \textcolor{black}{$35.6$ ($\textbf{+4.4}$) AP$^{\text{mask}}_{\text{novel}}$} with a shorter \textcolor{black}{training epochs}}.

\subsection{Ablation Study}
\begin{figure*}[t]
    \centering
    \includegraphics[width=1\linewidth]{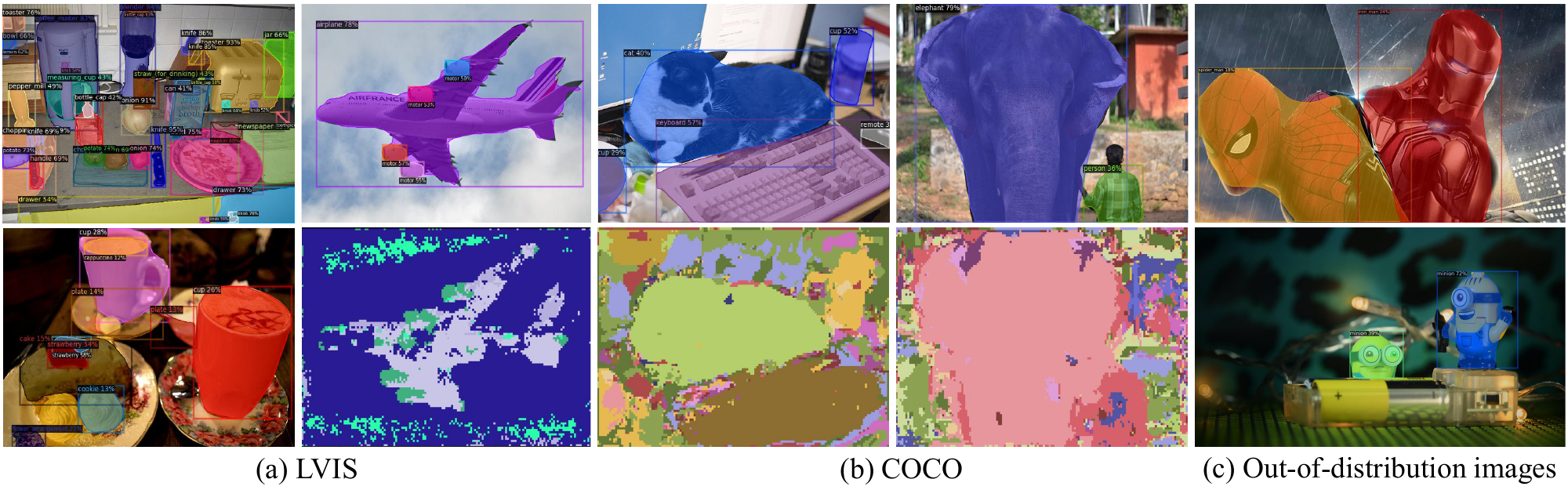}
    \vspace{-6mm}
    \caption{
     \textbf{Qualitative results of EdaDet.} We visualize our detection results and semantic maps on (a) LVIS~\cite{lvis}, (b) COCO~\cite{coco} and (c) out-of-distribution images from the open-source website. 
     }
    \label{fig:outof}
    \vspace{-6mm}
\end{figure*}

\noindent\textbf{Roadmap to build a strong open-vocabulary detector} is shown in Table~\ref{tab:ablations}. 
We start by training a conventional deformable DETR~\cite{zhu2020deformable} detector on base categories, which achieves $61.7$ AP50 of base categories. 
(1) To equip the base model with open-vocabulary detection ability, we replace its classifier with text embeddings of categories extracted from CLIP-R50. (2) Then, we deeply decouple the proposal generation and classification branches. To train the proposal generation branch on more objects beyond base objects in images, we use the annotated base objects and the generated proposals with high confidence scores and without overlapping with annotated objects as the box supervision. Note that we do not generate pseudo novel proposals from external training resources but only extend the base objects with class-agnostic proposals with high confidence scores. (3) Next, we ensemble text prompts to obtain a better classifier. So far, the model can achieve $15.1$ AP50 of novel categories, while the AP50 of base categories yields a performance drop by $-4.2$. 

Furthermore, we stepwisely develop our early dense alignment (Eda). (4) By replacing the object-level alignment with our dense-level alignment (4.1.1), the AP50 of novel categories significantly improves to $30.1$, demonstrating the effectiveness of preserving the fine-grained details to distinguish the similar novel and base objects. (5) We further add the mask $\mathcal{M}_k$ to allow non-object local image semantics in the object box to not directly participate in computing object categories (4.1.2 and 4.1.3), which helps mitigate overfitting base categories. The AP50 of novel categories is increased to $33.1$. (6) Moreover, we integrate CLIP's probability map in our dense probability map (4.1.4), slightly improving the AP50 of novel categories by $2.6$ while maintaining the AP50 of base categories to $57.2$. (7) We plug the global alignment into Eda (4.2), and the performance gain is $\mathbf{+2.1}$ AP50 on novel categories. 
In summary, our Eda, including the dense and global alignments, brings in $\mathbf{+22.7}$ improvement in AP50 of novel categories, which illustrates its effectiveness.

\begin{table}[t]
    \centering
    \small
    \begin{tabular}{lcc}
    
    \toprule
          \textbf{Ablation} & AP50$^{\text{box}}_{\text{base}}$ & AP50$^{\text{box}}_{\text{novel}}$  \\
    
    \midrule
    Supervised from base&        \color{gray}{$61.7$} &        $0.0$  \\
    ~(1)~Replace classifier with CLIP-R50 &  \color{gray}{$56.6$} & $7.6$ \\
    ~(2)~Deeply decouple from (1) &     \color{gray}{$57.3$} &         $14.2$    \\
    
    ~(3)~Add prompt ensembling from (2)    &        \color{gray}{$57.5$} &         $15.1$  \\
    
    ~(4)~Early Dense Alignment from (3)&     &   \\
    
    ~~(4.1)~Dense Alignment      &         &         \\
    ~~~(4.1.1)~$\lambda$ = 0, w/o $\mathcal{M}_k$       &        \color{gray}{$56.7$} &         $30.1$ \\
    ~~~(4.1.2)~$\lambda$ = 0, w $\mathcal{M}_{k={7 \times 7}}$       &        \color{gray}{$55.5$} &         $29.3$ \\
    ~~~(4.1.3)~$\lambda$ = 0, w $\mathcal{M}_{k={12 \times 12}}$       &        \color{gray}{$57.3$} &         $33.1$ \\
    ~~~(4.1.4)~$\lambda$ = 0.25, w $\mathcal{M}_{k={12 \times 12}}$       &        \color{gray}{$57.2$} &         $35.7$ \\
    
    ~~(4.2)~Global Alignment from (4.1.4)   & \color{gray}{$57.7$}   &    $37.8$      \\
    ~~(4.3)~Early Dense Alignment in con2\_x   & \color{gray}{$56.4$}   &    $35.9$      \\
    ~~(4.4)~Early Dense Alignment in con3\_x   & \color{gray}{$56.7$}   &    $36.2$      \\
    ~(5)~Self-training from (4.2)  & \color{gray}{$57.1$}   &    $40.2$  \\
    \bottomrule
    \end{tabular}
    \caption{\textbf{Ablation study of EdaDet} on the COCO dataset. }
    \label{tab:ablations}
\vspace{-16pt}
\end{table}

\begin{table}[t]
\small
\begin{center}
\begin{tabular}{lcccc}

\toprule
\multirow{1}{*}{Methods} & \multicolumn{1}{c}{AR} & \multicolumn{1}{c}{AR$_{\text{S}}$} & \multicolumn{1}{c}{AR$_\text{M}$} & \multicolumn{1}{c}{AR$_{\text{L}}$} \\ [.1em]
\shline

Mask-RCNN  & \phantom{1}0.31 & \phantom{1}0.19 & \phantom{1}0.39 & \phantom{1}0.53 \\

Def-Detr  & \phantom{1}0.33 & \phantom{1}0.21 & \phantom{1}0.41 & \phantom{1}0.51 \\

Decomp	&	\phantom{1}0.36	&	\phantom{1}0.24	&	\phantom{1}0.46	&	\phantom{1}0.57 \\

\hline
\end{tabular}
\end{center}
\vspace{-16pt}
\caption{\textbf{Ablation study of proposal generation.} We report top-$300$ average recall (AR) of all categories on LVIS. 
The AR$_\text{S}$, AR$_\text{M}$ and AR$_\text{L}$ denote AR for the small, medium and large objects, respectively. 
All methods are trained on LVIS base categories.}
\vspace{-16pt}
\label{tab:proposal}
\end{table}

\noindent\textbf{How much does deeply decoupling of two branches help proposal generation}.
Compared to Deformable DETR~\cite{zhu2020deformable} and Mask-RCNN~\cite{he2017mask}, the consistent improvement on all metrics indicates the superiority of deeply decoupling label prediction and box regression (see Table~\ref{tab:proposal}).

\subsection{Qualitative Visualization}

We visualize EdaDet’s detection results and semantic maps $\mathcal{S}_\text{detector}$ (Eq~\ref{equ:detector}) in Figure~\ref{fig:outof}. 
For LVIS with diverse target categories and complex scenes, our EdaDet performs well on crowded prediction. 
Figure~\ref{fig:outof}\textcolor{red}{c} shows that EdaDet even successfully detects the film characters \textit{iron man}, \textit{spider man} and the animation character \textit{minion}, which demonstrates EdaDet's open-vocabulary capacity and the importance of generalizing pretrained VLMs on open-vocabulary detection.

\section{Conclusion}
\vspace{-5pt}
We propose a simple but effective open-vocabulary detection method (EdaDet) that generalizes the pretrained VLMs to achieve a strong base-to-novel detection ability. Experiments on various datasets show that EdaDet consistently outperforms state-of-the-art methods in open-vocabulary object detection and instance segmentation. 
\noindent\textbf{Ethics Statement:}
Since our open-vocabulary capability is solely derived from VLMs, biases, stereotypes and controversies that may exist in the image-text pairs for training VLMs may be introduced into our models.

\noindent\textbf{Acknowledgment:} This work was supported by the National Natural Science Foundation of China (No.62206174), Shanghai Pujiang Program (No.21PJ1410900), Shanghai Frontiers Science Center of Human-centered Artificial Intelligence (ShangHAI), MoE Key Laboratory of Intelligent Perception and Human-Machine Collaboration (ShanghaiTech University), and Shanghai Engineering Research Center of Intelligent Vision and Imaging.

{\small
\bibliographystyle{ieee_fullname}
\bibliography{egbib}
}

\end{document}